%% file: template.tex
\documentclass{article}

\usepackage{arxiv}

\usepackage[utf8]{inputenc} 
\usepackage[T1]{fontenc}    
\usepackage{hyperref}       
\usepackage{url}            
\usepackage{booktabs}       
\usepackage{amsfonts}       
\usepackage{nicefrac}       
\usepackage{microtype}      
\usepackage{graphicx}
\usepackage[numbers]{natbib}

\usepackage{doi}
\usepackage{amsmath}
\usepackage{float}        
\usepackage{tabularx}     
\usepackage{multirow}     
\usepackage{booktabs}     
\usepackage{array}        

\usepackage{tabularx}
\newcolumntype{C}{>{\centering\arraybackslash}X}

\usepackage[ruled,vlined]{algorithm2e}

\title{Correctness Coverage Evaluation for Medical Multiple-Choice Question Answering Based on the Enhanced Conformal Prediction Framework}

\author{
Yusong Ke$^{1}$ \quad
Hongru Lin$^{2}$ \quad
Yuting Ruan$^{3}$ \quad
Junya Tang$^{1}$\thanks{Corresponding Author} \quad
Li Li$^{1}$ \\
$^{1}$School of Electronic and Information Engineering, Tongji University \\
$^{2}$School of Electrical and Information Engineering, Hunan University \\
$^{3}$School of Economics and Management, Fuzhou University \\
\texttt{\{2253737, junyatang, lili\}@tongji.edu.cn} \\
\texttt{2625277644@hnu.edu.cn} \quad \texttt{072203234@fzu.edu.cn}
}

\renewcommand{\shorttitle}{Correctness Coverage for Medical MCQA}

\hypersetup{
  pdftitle={Correctness Coverage Evaluation for Medical Multiple-Choice Question Answering Based on the Enhanced Conformal Prediction Framework},
  pdfauthor={Yusong Ke, Hongru Lin, Yuting Ruan, Junya Tang, Li Li},
  pdfkeywords={Large Language Models, Conformal Prediction, Medical Multiple-Choice Question-Answering, Average Prediction Set Size},
}

\begin{document}

\maketitle

\begin{abstract}
Large language models (LLMs) are increasingly adopted in medical question-answering (QA) scenarios. However, LLMs can generate hallucinations and nonfactual information, undermining their trustworthiness in high-stakes medical tasks. Conformal Prediction (CP) provides a statistically rigorous framework for marginal (average) coverage guarantees but has limited exploration in medical QA. This paper proposes an enhanced CP framework for medical multiple-choice question-answering (MCQA) tasks. By associating the non-conformance score with the frequency score of correct options and leveraging self-consistency, the framework addresses internal model opacity and incorporates a risk control strategy with a monotonic loss function. Evaluated on MedMCQA, MedQA, and MMLU datasets using four off-the-shelf LLMs, the proposed method meets specified error rate guarantees while reducing average prediction set size with increased risk level, offering a promising uncertainty evaluation metric for LLMs.
\end{abstract}

\keywords{Large Language Models \and Conformal Prediction \and Medical Multiple-Choice Question-Answering \and Average Prediction Set Size}


\section{Introduction}

Recently, large language models (LLMs)\cite{zhang2023spot,bi2025prism,bi2024visual,yang2025selfgoal,zhao2025chartcoder} have gained prominence as revolutionary instruments in real-world medical question-answering (QA) applications\cite{singhal2025toward,yang2025lighthouse}, offering significant potential to help both patients and healthcare professionals effectively address medical inquiries~\cite{thirunavukarasu2023large,brown2020language,hager2024evaluation}. In contemporary healthcare practices, many people rely on online search engines to obtain health-related information~\cite{he2025survey}. Despite the promising capabilities of LLMs, concerns persist about the reliability of their generated responses~\cite{das2025security,huang2025survey,yang2024logu}, particularly in light of the potential hazards associated with inaccurate or contradictory information. Such risks are especially critical in clinical settings, where misinformation can lead to adverse outcomes and potentially endanger patient safety~\cite{ouyang2022training}. Unlike general-domain QA tasks, medical QA often involves domain-specific terminology, context-sensitive reasoning, and high-stakes decision-making, where even minor inaccuracies may have significant consequences. Therefore, ensuring the reliability and trustworthiness of model outputs is not merely a performance issue, but a fundamental requirement for real-world applicability. These challenges highlight the need for rigorous uncertainty quantification frameworks with formal guarantees, and strongly motivate the adaptation and extension of conformal prediction methods specifically for the medical QA domain.

Uncertainty quantification (UQ) has emerged as a critical methodology for assessing the reliability of LLM output~\cite{kendall2017uncertainties,kadavath2022language,xiong2023can,chen2024quantifying,shorinwa2024survey}. Existing UQ techniques can be broadly categorized into white-box and black-box approaches. White-box methods utilize internal model information, such as logarithmic summation, to quantify semantic uncertainty~\cite{kuhn2023semantic,duan-etal-2024-shifting,wang2025word}, including metrics such as semantic entropy~\cite{farquhar2024detecting}. In contrast, black-box uncertainty measures, rooted in self-consistency theory, evaluate uncertainty based on the semantic diversity of multiple outputs generated for the same input without requiring access to the model's internal states~\cite{lin2023generating,wang2024conu,qiu2024semantic}. Despite their utility, existing methods rely on heuristic definitions of uncertainty and lack of rigorous statistical guarantees for task-specific performance metrics~\cite{jin2023selection}.

Conformal prediction (CP) is now recognized as a robust framework within the broader domain of machine learning, offering statistically rigorous guarantees of marginal (average) coverage for prediction sets~\cite{angelopoulos2021gentle,campos2024conformal}. Previous research has demonstrated the successful adaptation of CP to classification tasks~\cite{angelopoulos2021uncertainty,angelopoulos2024theoretical,huang2024confine}, achieving error correction miscoverage rates defined by the user. However, extending CP to medical question-answering (QA) tasks presents substantial challenges. 

To address these limitations, this study proposed an enhanced CP framework, which develops the non-conformity score (NS) to be closely aligned with the correct option in multiple-choice medical question-answering (MCQA) tasks. Given that access to internal model information may not be feasible in practical medical MCQA scenarios, the enhanced CP framework defines the NS as one minus the probability of the correct option for each calibration data point, generates multiple outputs for the same medical query, and computes the frequency score for each option. Using self-consistency theory~\cite{wang2022self}, the enhanced CP framework utilizes the frequency score within the sampling set as a proxy for the probability of the option.
Furthermore, to address the limitation that the enhanced CP framework only provides rigorous control over the correct answer coverage, this study presents a risk control framework to manage task-specific metrics, ensuring that equivalent results can be obtained when the loss function is designed to reflect the correctness error miscoverage. So that equivalent results can be obtained when the loss function is designed to reflect the correctness error miscoverage 

The proposed enhanced CP framework was evaluated on three multiple medical MCQA datasets, including the MedMCQA~\cite{jin2021disease}, MedQA~\cite{jin2021disease}, and MMLU~\cite{hendryckstest2021} datasets,  utilizing four popular pre-trained LLMs, including Llama-3.2-1B-Instruct, Llama-3.2-3B-Instruct, Qwen2.5-1.5B-Instruct, and Qwen2.5-3B-Instruct. 
Empirical results demonstrate that the enhanced CP framework achieves strict control over the marginal error rate (average correctness miscoverage rate) on test MCQA samples at various risk levels. 
Moreover, the average prediction set size (APSS) within the test set is also estimated in this study.
Experimental findings show that the APSS metric decreases as the risk level increases, which can assist in using indicators such as accuracy to comprehensively evaluate the performance of LLMs. 
The main contributions of this study are summarized as follows. (i) For the first time, conformal prediction is applied to the medical multiple-choice question-answering tasks, strictly guaranteeing the coverage of correct options. (ii) A promising metric is revealed to assess the uncertainty of LLM by estimating the average prediction set size on the test MCQA sample. (iii) Extensive evaluation and ablation studies on three popular medical benchmarks are conducted using four pre-trained LLMs, demonstrating the effectiveness and robustness of the enhanced CP framework.

The remainder of this study is organized as follows. Section 2 introduces related works. In section 3, a detailed enhanced CP framework has been illustrated and explained. An experiment is presented in section 4. Section 5 concludes this study and gives future works.


\section{Related Work}
Medical question-answering (QA) presents distinct challenges compared to general-domain QA tasks, owing to its reliance on domain-specific terminology, the necessity for precise multi-hop reasoning across heterogeneous medical knowledge sources, and the potentially severe consequences of erroneous outputs. Furthermore, many medical queries are inherently ambiguous or highly context-dependent, and the corresponding data are often incomplete or uncertain.

These complexities hinder the ability of conventional QA models to produce reliable responses, particularly in open-ended or generative settings. To mitigate these limitations, recent research has introduced self-consistency-based approaches, which enhance model robustness by measuring the agreement across multiple independently generated responses. Building upon this paradigm, the present study integrates self-consistency mechanisms within a conformal prediction framework, enabling the construction of uncertainty-aware prediction sets that offer formal statistical coverage guarantees, while maintaining task-specific reliability in high-stakes medical QA scenarios.

\subsection{Natural Language Generation tasks in the medical domain}
The application of Natural Language Generation (NLG) in the medical domain has become a critical area of research, driven by the growing reliance on advanced technologies to optimize patient care and clinical communication. NLG systems are increasingly employed to automate clinical documentation, address medical inquiries, and provide diagnostic suggestions. These systems utilize state-of-the-art large language models (LLMs) such as GPT-3~\cite{brown2020language} and BERT~\cite{devlin2019bert} to produce responses that are both contextually relevant and medically accurate. By delivering timely and precise information, these technologies aim to alleviate healthcare professionals' workload and enhance patient outcomes~\cite{ouyang2022training}.

While NLG models present considerable potential within medical applications, they also encounter substantial challenges. A primary concern is the reliability of the generated content, especially in high-stakes scenarios such as medical diagnosis and patient care. Despite their ability to develop contextually appropriate and persuasive responses, these models may also produce factually incorrect or misleading information, posing significant risks in clinical environments. Recent research efforts have focused on implementing domain-specific fine-tuning strategies and integrating uncertainty quantification (UQ) techniques to assess and enhance the reliability of NLG systems in medical contexts~\cite{lin2023generating}. Nonetheless, further advancements are needed to ensure that NLG systems consistently meet the rigorous safety and accuracy standards of healthcare settings.

\subsection{Uncertainty Quantification in Medical Question-Answering Tasks}
Uncertainty Quantification (UQ) plays a pivotal role in medical question-answering (QA) tasks, where the consequences of inaccurate or unreliable responses can lead to severe clinical implications. In such high-stakes applications, it is imperative to rigorously assess the confidence of the system's predictions to determine when a generated answer can be considered trustworthy. Traditional UQ methodologies, predominantly designed for classification tasks, are not readily applicable to the inherently open-ended nature of medical QA, where responses often involve complex, nuanced, and context-dependent information.

Recent advancements in the field have introduced specialized UQ techniques tailored to the unique demands of medical QA tasks. Confidence-based metrics~\cite{kendall2017uncertainties} and entropy-based approaches~\cite{malinin2021uncertainty} have been employed to evaluate the uncertainty associated with medical responses. These methods typically involve analyzing the model's output distribution or assessing the consistency of predictions across multiple generated responses. Additionally, self-consistency frameworks and ensemble methods have been proposed to compute confidence scores by aggregating the outputs from diverse model iterations~\cite{lin2023generating}. Despite their potential, many existing approaches lack formal statistical guarantees, representing a significant limitation in clinical settings where precision and reliability are critical. Consequently, developing robust UQ methodologies is necessary to effectively manage the complexities of medical data while delivering reliable and interpretable uncertainty estimates.

\subsection{Conformal Prediction for Uncertainty Quantification in Medical Question-Answering Tasks}
Conformal Prediction (CP) has emerged as a robust methodology for uncertainty quantification (UQ) in medical question-answering (QA) tasks, providing a rigorous framework for constructing prediction sets with statistically guaranteed coverage~\cite{campos2024conformal,wang2025sconu}. Unlike traditional UQ techniques that produce single-point predictions, CP generates a set of potential outcomes alongside confidence measures, offering a quantifiable assessment of the reliability of a generated response. The capability is particularly critical in medical QA applications, where the implications of erroneous predictions can be severe, and clinicians must often rely on automated responses to inform clinical decision-making.

The utility of CP is further enhanced in the context of large language models (LLMs)\cite{ye2024benchmarking,wang2025sample,wang2024conu,Gui2024ConformalAK,angelopoulos2024theoretical}, which are frequently characterized as “black-box” models due to their inherent opacity~\cite{lin2023generating}. A key advantage of CP is its model-agnostic nature, as it does not require assumptions about the underlying data distribution, allowing its application across a broad spectrum of machine learning models, including LLMs. Recent studies have explored the adaptation of CP methodologies to open-ended natural language generation (NLG) tasks, such as medical QA, demonstrating the potential of CP to balance prediction accuracy with coverage. This controlled trade-off is particularly valuable in high-stakes domains like healthcare, where maintaining a balance between model performance and reliability is essential. Nevertheless, significant challenges remain in scaling CP approaches and managing the complexities of diverse and heterogeneous medical data.

\section{Method}
The methods presented in this section introduce a novel framework based on Conformal Prediction (CP) for medical multiple-choice question-answering tasks. This framework provides an advanced approach for uncertainty quantification (UQ) and task-specific risk management. By leveraging the framework, prediction sets are generated that are statistically guaranteed to contain the true label with a predefined confidence level, which is essential for applications in high-stakes fields such as healthcare.

\subsection{Adaptation of Conformal Prediction to Medical MCQA Tasks}

Conformal Prediction (CP) is a statistical framework that generates prediction sets with robust uncertainty quantification for machine learning models. Unlike traditional methods that provide single-point predictions, CP constructs a set of potential outcomes for each test instance, ensuring that the true label is included with at least \(1 - \alpha\) confidence. This characteristic is particularly valuable for high-stakes applications such as medical MCQA tasks.

\begin{figure*}[h]
    \centering
    \includegraphics[width=\linewidth]{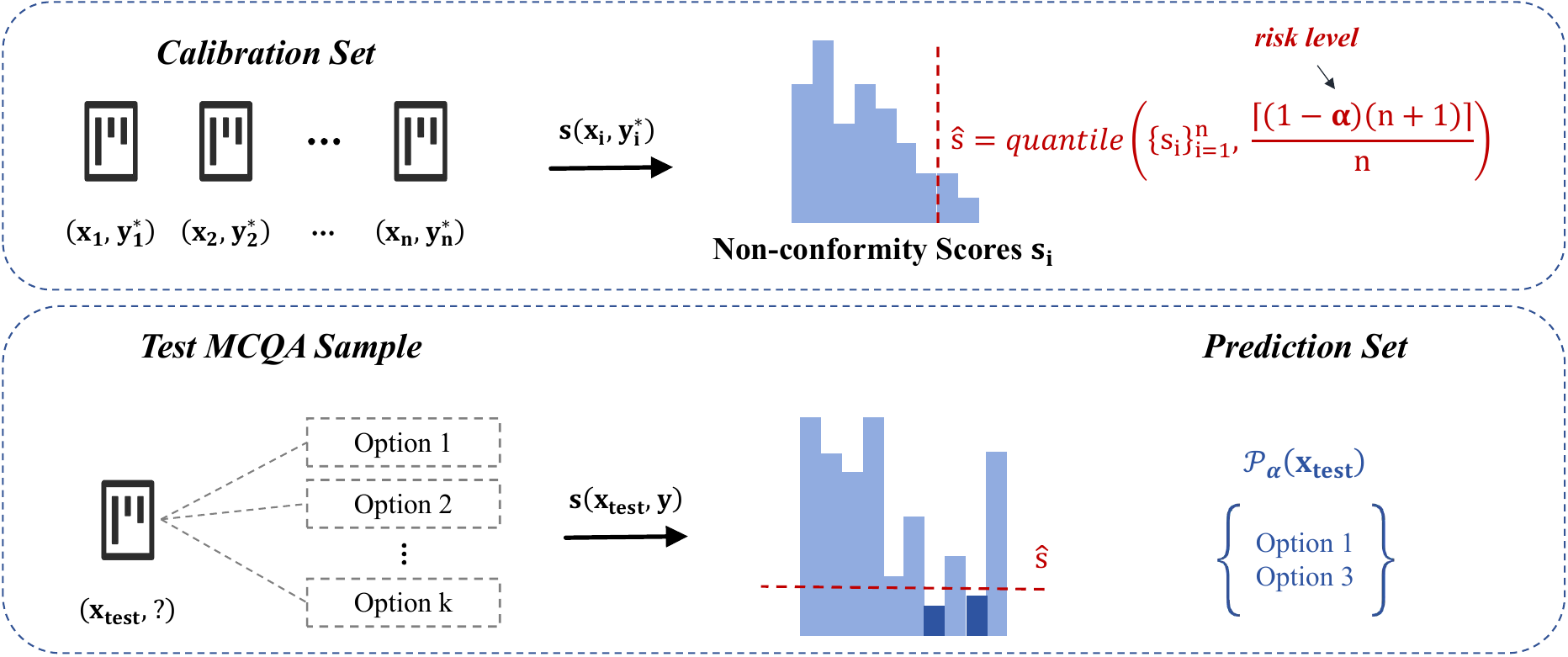}
    \caption{Adaptation of Conformal Prediction to Medical MCQA Tasks.}
    \label{fig:enter-label}
\end{figure*}

As illustrated in Figure~\ref{fig:enter-label}, the framework first measures the disagreement between the input \(x_i\) and its ground-truth label \(y_i^*\) by computing a non-conformity score for each calibration sample. A quantile threshold \(\hat{q}\) is then determined based on these scores to establish the statistical risk level. Given a test MCQA sample \(x_{\text{test}}\), the model evaluates multiple candidate answers, computing their respective non-conformity scores. The prediction set is subsequently constructed by retaining only those answer choices whose scores fall below the calibrated threshold, ensuring that the final prediction set adheres to the desired miscoverage rate guarantees.

In medical MCQA, CP calculates a conformity score \(s(X_{\text{test}}, Y_{\text{test}})\) for each test instance, comparing it with the empirical distribution of conformity scores from the training data. A threshold \( \hat{q} \) is then derived to define the prediction set \( C(X_{\text{test}}) \) that meets the desired coverage probability:

\begin{equation}
    C(X_{\text{test}}) = \{ y : s(X, y) \leq \hat{q} \}.
\end{equation}

The method ensures a statistically rigorous coverage guarantee:

\begin{equation}\label{eq:2}
    P(Y_{\text{test}} \in C(X_{\text{test}})) \geq 1 - \alpha.
\end{equation}

The calibration process further enhances this reliability by fine-tuning the threshold using a dedicated calibration set. However, while CP provides a robust approach to managing uncertainty in predictive models and offers well-calibrated prediction sets suitable for the stringent requirements of medical MCQA tasks, its limitation lies in the inability to manage task-specific metrics beyond mere coverage control. This shortcoming underscores the necessity of the proposed risk control framework, which extends the capabilities of CP by introducing a monotonic decreasing loss function to manage task-specific risks effectively and enhance the reliability of large language models in high-stakes medical applications.

\subsection{Guarantee of Task-specific Metrics}
This section introduces a method for constructing prediction sets with statistically rigorous guarantees on miscoverage rates via conformal calibration, specifically tailored for the medical multiple-choice question answering (MCQA) context. Unlike general-purpose QA tasks, medical MCQA involves domain-specific, often rare and compound terminology, which is typically segmented into multiple sub-tokens during language model processing. This characteristic introduces inconsistency in token-level semantic evaluations, leading to unstable and biased uncertainty estimates. To address this challenge, the proposed framework operates at the answer option level, treating each candidate response as a discrete class and computing non-conformity scores to quantify its deviation from correctness. By leveraging a calibration  set \(\{(x_i, y_i^*)\}_{i=1}^n\) and a test data point \((x_{n+1})\) , the method enables precise task-specific risk control through the design of customized loss functions, facilitating robust uncertainty quantification in high-stakes medical scenarios.

The proposed method extends traditional classification tasks into the Question Answering (QA) domain by leveraging the flexibility of the Non-Conformity Score. In a typical classification task, a model predicts a discrete label from a predefined set of classes. The classification task is transformed into a QA task by mapping each class label to a potential answer in a natural language format. This mapping is particularly effective when the model generates candidate answers requiring more contextual understanding and natural language generation.

For example, a classification task that categorizes images into “cat,” “dog,” or “bird” can be rephrased as a QA task by asking, “What animal is in the image?” The candidate answers become the possible labels, and the model evaluates each candidate's reliability through the Non-Conformity Score, ensuring that the prediction set adheres to the desired coverage guarantees.

Given a pre-trained language model \(f\), the input query \(x_i\) is utilized to generate a candidate set \(\{y_j^i\}_{j=1}^m\). For each candidate set, the prediction set \(P_i(t)\) is defined as:

\begin{equation}
    P_i(t) = \left\{ y \in \{y_j^i\}_{j=1}^m : R(y) \geq 1 - t \right\},
\end{equation}

\(R(y)\) is a quantitative measure of the reliability of a candidate's answer \(y\) within the set. The parameter \(t\) functions as a threshold, systematically regulating the inclusion of predictions based on their reliability scores.

To enable task-specific risk management, we introduce the Non-Conformity Score \(S(y, x_i)\), which quantifies the degree of deviation of a candidate prediction \(y\) from the expected true answer given the input \(x_i\). The Non-Conformity Score is defined as:

\begin{equation}
    S(y, x_i) = 1 - R(y),
\end{equation}

\(R(y)\) represents the reliability score of the candidate answer. The Non-Conformity Score is critical in bridging classification and QA tasks by providing a unified metric for evaluating prediction quality across different task types.

In the classification setting, \(S(y, x_i)\) can be interpreted as a measure of how likely a predicted label is to be incorrect. When extended to QA tasks, this score helps determine the most suitable candidate answers by evaluating their conformity with the context of the question. By setting appropriate thresholds, the method ensures that only highly reliable answers are included in the prediction set, effectively managing task-specific risks.

For each calibration data point, the loss function \(l_i(t)\) is defined as:

\begin{equation}
    l_i(t) = 1\{y_i^* \notin P_i(t)\},
\end{equation}

\(y_i^*\) is the ground-truth answer, the loss function is designed as a non-increasing function with respect to the tuning parameter \(t\), ensuring that a higher parameter value does not increase the loss. This property is critical for maintaining the theoretical monotonicity required for robust empirical risk management.

The empirical loss \(L_n(t)\) over the calibration set is defined as:

\begin{equation}
    L_n(t) = \frac{1}{n}\sum_{i=1}^n l_i(t).
\end{equation}

The primary objective of the proposed method is to control the expected loss of new test data by ensuring:

\begin{equation}\label{eq:7}
    E[L_{n+1}(t)] \leq \alpha.
\end{equation}

Under the exchangeability assumption of calibration and test data, the optimal parameter \(\hat{t}\) is derived by solving the following optimization problem:

\begin{equation}
    \hat{t} = \inf \left\{ t : L_n(t) \leq \frac{\alpha(n+1)-1}{n} \right\}.
\end{equation}

This formulation guarantees a statistically rigorous control of the miscoverage rate, thereby offering robust theoretical assurances for predictive accuracy and reliability. Beyond ensuring marginal coverage, the proposed method provides flexible support for task-specific metric control through a customized loss function design. By tailoring the loss function to the characteristics of classification and question-answering (QA) tasks, this framework presents a highly adaptable risk management strategy. Integrating the non-conformity score further enhances the model’s ability to maintain stable uncertainty control across various prediction scenarios, which is particularly beneficial in high-stakes applications requiring strong statistical guarantees.

The complete procedure is summarized in Algorithm~\ref{alg:confine}, which details each step in constructing prediction sets with conformal guarantees for medical MCQA tasks, and demonstrates how statistical rigor is operationalized into a practical inference framework.

\begin{algorithm}[!h]
\caption{Frequency-Based Implementation of Conformal Prediction for Medical MCQA}\label{alg:confine}
\LinesNumbered

\KwIn{
    Calibrated sampled generations $\mathcal{D}^c = \left\{ (x_i, \{y_j^i\}, y_i^*) \right\}_{i=1}^m$; \\
    Test sample $x_{m+1}$ with sampled responses $\{y_j^{(m+1)}\}$; \\
    {User-specified significance level $\varepsilon$}; Class number $C$.
}

\ForEach{$i \in \{1, \dots, m\}$}{
    Count the occurrences of each option $A$ to $D$ in $\{y_j^i\}$ to estimate empirical probabilities $\hat{p}_{i,1}, \dots, \hat{p}_{i,C}$
}

\ForEach{$i \in \{1, \dots, m\}$}{
    Let $y_i^*$ be the correct label for $x_i$ \\
    Compute non-conformity score: $\alpha_i = 1 - \hat{p}_{i, y_i^*}$
}

Count class frequencies in $\{y_j^{(m+1)}\}$ to obtain $\hat{p}_{m+1,1}, \dots, \hat{p}_{m+1,C}$

\For{$j \leftarrow 1$ \KwTo $C$}{
    $\alpha_j^{(m+1)} = 1 - \hat{p}_{m+1,j}$ \\
    $p\left(Y_j^{(m+1)}\right) = \frac{1}{m+1} \cdot \#\left\{i : \alpha_i \geq \alpha_j^{(m+1)}\right\}$ \\
    \If{$p\left(Y_j^{(m+1)}\right) > \varepsilon$}{
        Add $Y_j^{(m+1)}$ to prediction set $\Gamma^\varepsilon$
    }
}

Prediction $\leftarrow \arg\max_j \hat{p}_{m+1,j}$

\KwOut{Prediction, $\Gamma^\varepsilon$}
\end{algorithm}

\section{Experiments}
The experiments presented in this section aim to evaluate the performance of the proposed framework, utilizing a set of widely recognized medical datasets and state-of-the-art large language models (LLMs). These experiments focus on assessing the framework's ability to manage uncertainty quantification (UQ) and task-specific risks across various medical question-answering (QA) tasks. Additionally, the performance of the framework is measured using several standard metrics, providing a comprehensive analysis of its effectiveness in real-world applications.
\subsection{\textit{Experimental Settings}}
\subsubsection{\textit{Datasets}}

The study utilizes three widely recognized benchmark datasets—MedMCQA, MedQA, and the multi-task MMLU dataset—to rigorously assess the performance of the proposed approach across diverse medical question-answering (QA) scenarios. To ensure a robust and unbiased evaluation, we uniformly sampled 2,000 representative instances from each dataset to maintain consistency and comparability throughout the experimental design.

The selected datasets provide a broad and diverse benchmark for evaluating medical and general knowledge QA systems. The MedMCQA dataset is curated explicitly for real-world medical entrance examinations, encompassing over 194,000 high-quality multiple-choice questions~\cite{jin2021disease}. It spans 2,400 healthcare topics across 21 medical subjects, requiring advanced reasoning and analytical skills. The MedQA dataset offers a multilingual multiple-choice question dataset. It includes 12,723 questions in English, 34,251 in simplified Chinese, and 14,123 in traditional Chinese, along with a large-scale corpus of medical textbook content to support reading comprehension models.

The MMLU dataset evaluates multi-task language understanding across 57 diverse tasks, including humanities, social sciences, hard sciences, and law. It challenges models with both general knowledge queries and advanced problem-solving scenarios. Together, these datasets serve as robust and diverse evaluation tools, enabling a thorough assessment of model performance in specialized medical and broad general knowledge domains, and facilitating the advancement of more reliable and adaptable QA systems.

\subsubsection{\textit{Based LLMs}}
This work evaluates the proposed framework using four large language models (LLMs), each with distinct architectures and model sizes.
 Specifically, the models employed include Llama-3.2-1B-Instruct~\cite{touvron2023llama}, Llama-3.2-3B-Instruct~\cite{touvron2023llama}, Qwen2.5-1.5B-Instruct~\cite{yang2024qwen2}, and Qwen2.5-3B-Instruct~\cite{yang2024qwen2}. These models span a parameter range from 1.5 billion to 3 billion, enabling a comprehensive analysis of how model size influences performance across diverse tasks.

The Llama-3.2-1B-Instruct~\cite{touvron2023llama} and Llama-3.2-3B-Instruct~\cite{touvron2023llama} models are built upon the Llama architecture, which is explicitly designed to excel in instruction-following scenarios. These models are particularly effective in tasks requiring a nuanced understanding of natural language instructions and generating coherent and contextually appropriate responses. Conversely, the Qwen2.5-1.5B-Instruct~\cite{yang2024qwen2} and Qwen2.5-3B-Instruct~\cite{yang2024qwen2} models are optimized for multi-task learning, performing effectively in various NLP applications, including classification, reasoning, and responding to questions. By incorporating LLMs with diverse architectures and parameter scales, the proposed experimental design enhances the robustness and generalizability of the evaluation. This approach ensures a thorough and nuanced assessment of the proposed framework's effectiveness across models of varying sizes and functional capabilities, contributing to a more holistic understanding of its performance dynamics.

\subsubsection{\textit{Hyperparameters}}

In the experiments, the split ratio between the calibration and test sets is fixed at 0.5, ensuring an equal data distribution across both sets. This strategy provides a balanced and reliable evaluation of model performance. Consistent with prior studies, multinomial sampling is utilized to generate \(M\) candidate responses for each data point. For the MMLU and MedMCQA datasets, each comprising four answer options per question, the number of candidate responses \(M\) is set to 20, aligning with established practices~\cite{kuhn2023semantic}. For the MMLU-Pro dataset, where each sample includes ten multiple-choice options, \(M\) is increased to 50 to approximate the model output distribution more accurately.

To optimize computational efficiency in multiple-choice question-answering (MCQA) tasks, the maximum generation length is set to 1, and the generation temperature is fixed at 1.0 for sampling. A broader set of hyperparameters is explored for open-domain QA tasks, with further performance refinement achieved through risk calibration techniques as detailed in prior research~\cite{angelopoulos2021gentle,jin2023selection,huang2024confine}. 
The parameter \texttt{input\_length} refers to the embedding length of the input prompt after being encoded by the tokenizer of the current language model. Additionally, the maximum generation length for both tasks is set to 36 tokens, ensuring that generated responses are sufficiently detailed while avoiding excessive constraints imposed by model limitations.

\subsubsection{\textit{Evaluation Metrics}}

The performance is rigorously assessed using two principal metrics: the Empirical Miscoverage Rate (EMR) and the Average Prediction Set Size (APSS). The EMR assesses how well the proposed framework maintains the intended marginal coverage within the test set. It measures the frequency at which the true label of a test sample falls outside the predicted coverage set, offering a quantitative assessment of prediction reliability. A thorough analysis of the EMR verifies that the model consistently achieves the desired coverage level under various test conditions, demonstrating its robustness and generalization capabilities.

In addition to EMR, the APSS metric is utilized to assess prediction efficiency and uncertainty for each LLM. The APSS represents the average number of candidate responses generated per test sample, indicating the model's handling of uncertainty in the prediction process. A lower APSS reflects a more efficient model that produces fewer yet accurate responses, while a higher APSS may indicate increased uncertainty or a more conservative predictive approach. By combining EMR and APSS metrics, the proposed framework offers a comprehensive analysis of model performance, balancing predictive reliability with efficiency and providing valuable insights into its behavior in real-world applications.

\subsection{Empirical Evaluations}
 
\subsubsection{\textit{Statistical Guarantees of the EMR Metric}}

This section rigorously validates the calibrated prediction sets constructed according to Eq.~\ref{eq:2}, showing that they reliably achieve the desired coverage levels across different user-defined error rates. Furthermore, the practical effectiveness is investigated by applying selective prediction guided by the proposed uncertainty metric.

\subsubsection{\textit{Empirical Coverage Guarantees}}

\begin{figure*}[h]
    \centering
    \includegraphics[width=0.8\linewidth]{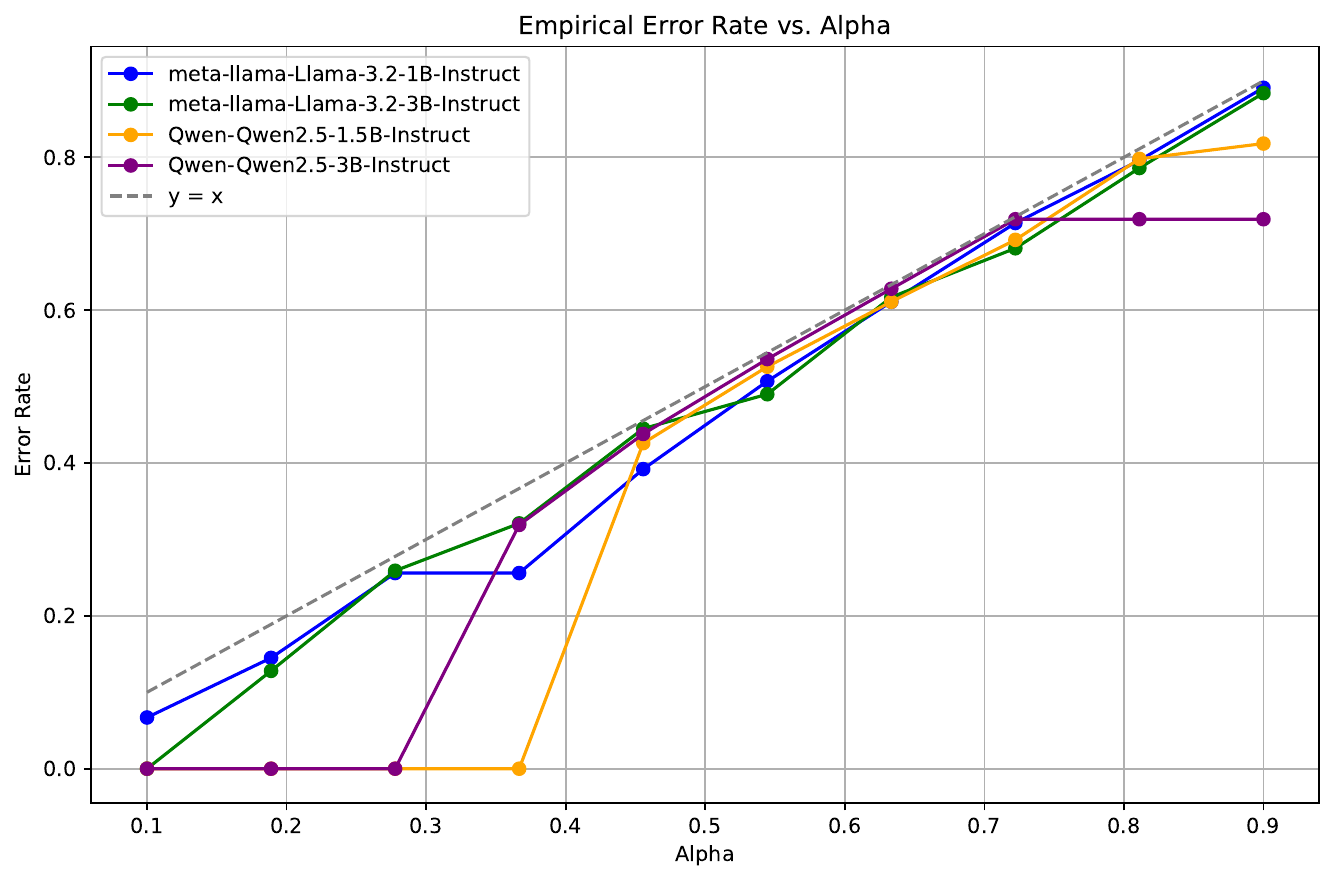}
    \caption{Empirical Miscoverage Rate (EMR) for the MedMCQA dataset across different confidence levels (\( \alpha \)). The Llama models exhibit superior stability and lower error rates compared to the Qwen2.5 models.}
    \label{fig:medmcqa_emr}
\end{figure*}

\begin{figure*}[h]
    \centering
    \includegraphics[width=0.8\linewidth]{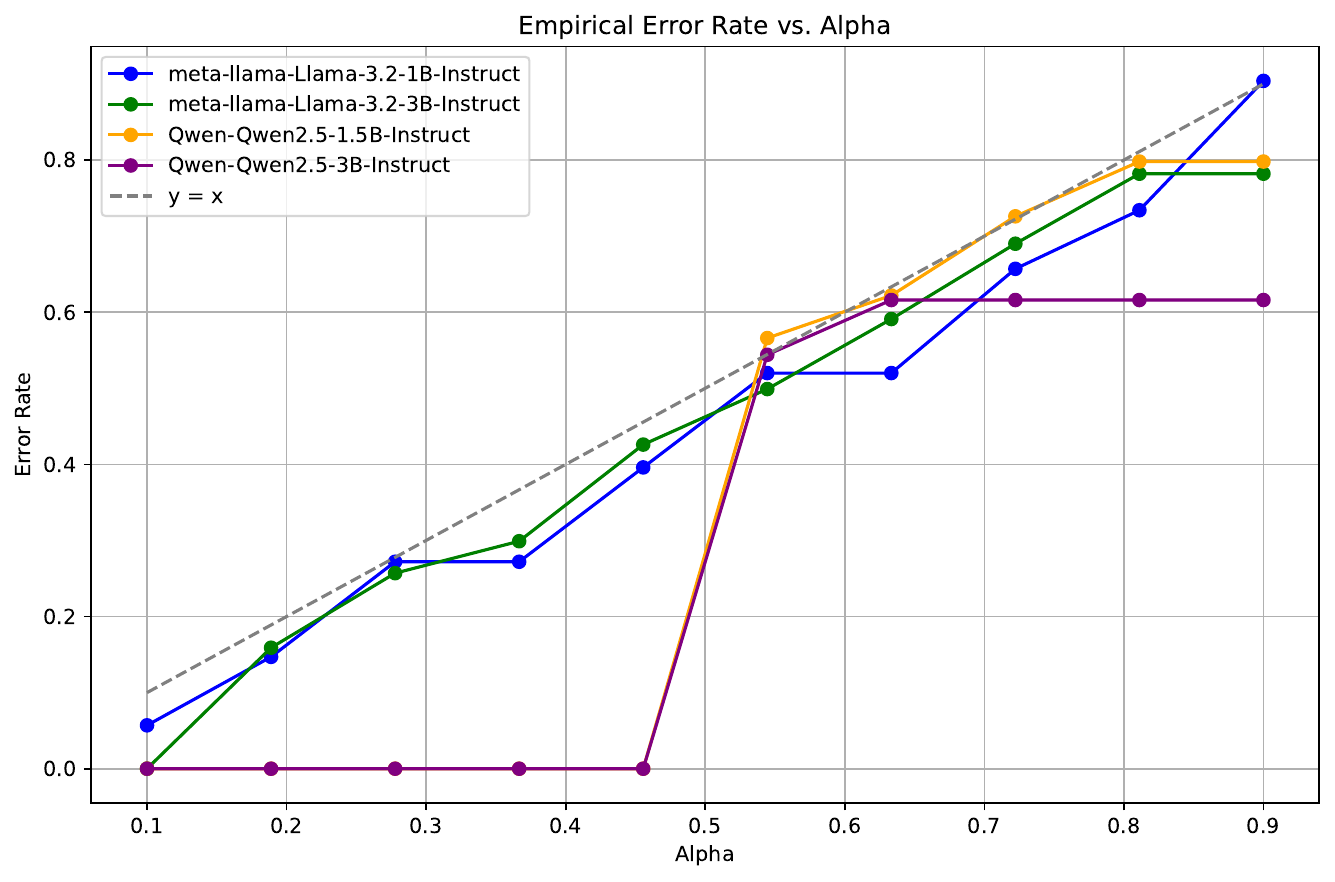}
    \caption{EMR performance on the MedQA dataset, demonstrating the consistent reliability of Llama models over a range of error rates, with the Qwen2.5 models showing various variability.}
    \label{fig:medqa_emr}
\end{figure*}

\begin{figure*}[h]
    \centering
    \includegraphics[width=\linewidth]{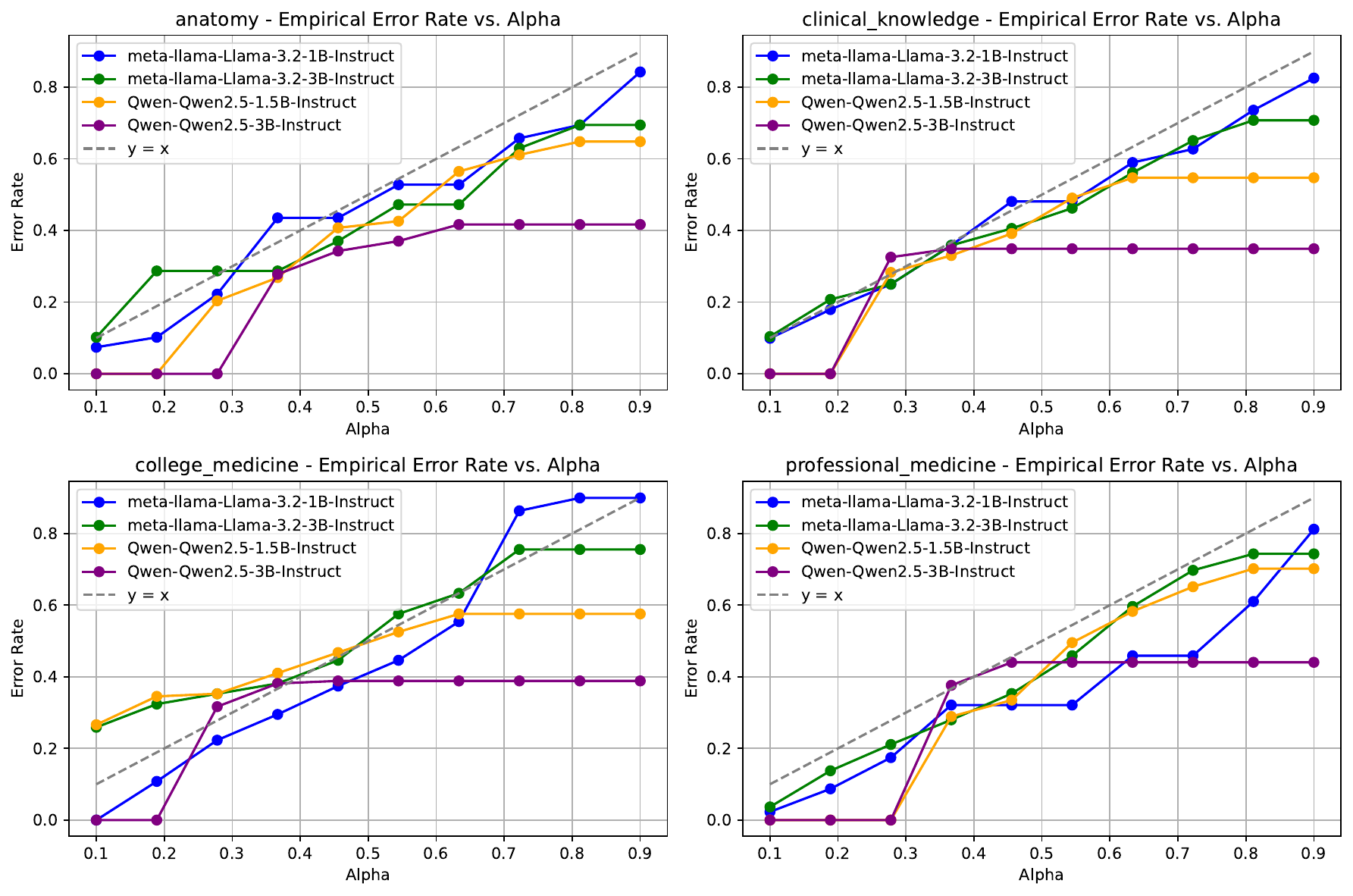}
    \caption{EMR results for the MMLU dataset across categories including high school biology, anatomy, clinical knowledge, and college medicine. The Llama models maintain better stability and accuracy, particularly in clinical knowledge tasks.}
    \label{fig:mmlu_emr}
\end{figure*}

To verify the minimum correctness coverage level empirically, each of the four datasets is partitioned with a 1:10 ratio into calibration and testing subsets. The calibration subset determines the conformal uncertainty threshold based on the maximum allowable error rate. Subsequently, coverage performance is assessed using the testing subset, and corresponding results for MedMCQA, MedQA, and MMLU are shown in Figures~\ref{fig:medmcqa_emr}, \ref{fig:medqa_emr}, and \ref{fig:mmlu_emr}.

In Figure~\ref{fig:medmcqa_emr}, which displays the MedMCQA dataset results, a consistent trend of increasing error rates is observed as the confidence level \( \alpha \) rises. Notably, the Llama-3.2-1B-Instruct~\cite{touvron2023llama} and Llama-3.2-3B-Instruct~\cite{touvron2023llama} models consistently maintain lower and more stable error rates compared to the Qwen2.5-1.5B-Instruct~\cite{yang2024qwen2} and Qwen2.5-3B-Instruct~\cite{yang2024qwen2} models, which exhibit more significant variability.

A similar pattern is evident in Figure~\ref{fig:medqa_emr}, representing the MedQA dataset. The Llama models again demonstrate superior stability and lower error rates, whereas the Qwen2.5 models show an increase in error rates at higher \( \alpha \) values. The findings indicates that the Llama models offer more reliable predictive performance across a broad range of confidence levels.

For the MMLU dataset, as shown in Figure~\ref{fig:mmlu_emr}, four distinct categories are analyzed: high school biology, anatomy, clinical knowledge, and college medicine. Across all categories, the Llama models consistently exhibit enhanced stability and lower error rates relative to the Qwen2.5 models. The performance discrepancy is mainly pronounced in the clinical knowledge category, where the Qwen2.5 models show significant fluctuations in error rates.

These empirical results validate the stringent control over the correctness coverage rate under varying error rate conditions across all datasets. The statistical guarantees provided by these findings align closely with the theoretical framework established in Eqs.~\ref{eq:2} and~\ref{eq:7}, demonstrating that the calibrated prediction sets constructed by the proposed method ensure robust and statistically rigorous correctness coverage under diverse calibration scenarios.

\subsubsection{\textit{Uncertainty Estimation of LLMs}}

The Average Prediction Set Size (APSS) metric is a critical indicator of model uncertainty, demonstrating a clear inverse relationship with the risk level \( \alpha \). As illustrated in Table~\ref{tb:apss_results}, the APSS consistently decreases as the risk level increases, reflecting a more stringent criterion for constructing prediction sets. This trend is evident across all evaluated models, underscoring the robustness of the APSS metric as a measure of predictive uncertainty.

For each dataset, an increase in the risk level \( \alpha \) generally reduces the APSS metric. The more larger models typically exhibit smaller average prediction set sizes than the Qwen2.5 series models. For instance, in the MedMCQA dataset, the Llama-3.2-1B-Instruct~\cite{touvron2023llama} model reduces its APSS from 3.569 at \( \alpha = 0.1 \) to 0.214 at \( \alpha = 0.9 \). Similarly, in the MedQA dataset, the Llama-3.2-3B-Instruct model demonstrates a pronounced decrease in APSS from 4.000 at \( \alpha = 0.1 \) to 0.137 at \( \alpha = 0.9 \), highlighting a consistent narrowing of the prediction sets as the uncertainty threshold tightens.

\input{Table/APSS_results}

In contrast, the Qwen2.5 models exhibit a more gradual reduction in APSS with increasing \( \alpha \) values. Specifically, the Qwen2.5-1.5B-Instruct model~\cite{yang2024qwen2} on the MMLU dataset maintains an APSS of 4.000 up to \( \alpha = 0.5 \), after which a gradual decrease is observed. This behavior suggests that the Qwen2.5 models tend to generate larger and more stable prediction sets across varying confidence levels, potentially indicating higher uncertainty in their predictions.

These empirical findings validate that the APSS metric effectively captures model uncertainty under different error thresholds. A lower APSS correlates with reduced uncertainty and a more precise prediction set, aligning well with theoretical expectations of selective prediction. As the model's confidence increases, the prediction set size appropriately contracts, demonstrating the model's capability to manage uncertainty. Additionally, the efficiency of different models is highlighted, with the larger Llama models demonstrating superior efficiency in narrowing prediction sets compared to the Qwen2.5 models, which maintain larger prediction sets even at higher \( \alpha \) levels.

Overall, the APSS metric serves as a reliable measure of model uncertainty and offers critical insights into model optimization strategies for achieving more precise and selective predictions under varying confidence thresholds. It reinforces the utility of APSS as a tool for evaluating model performance in high-stakes applications where prediction certainty is paramount.

\subsection{Sensitivity Analysis}
\subsubsection{\textit{EMR at Various Split Ratios}}

To assess the sensitivity of the proposed method to different data partitioning strategies, a comprehensive analysis of the  Empirical Miscoverage Rate (EMR) was conducted on the MedMCQA dataset, with the parameter \( \alpha = 0.1 \). The evaluation was performed across a range of split ratios (0.1, 0.3, 0.5, 0.7), as detailed in Table~\ref{tb:EMR_at_various_split_ratios}. The results demonstrate that the \textit{Llama-3.2-1B-Instruct}~\cite{touvron2023llama} and \textit{Llama-3.2-3B-Instruct}~\cite{touvron2023llama} models maintained consistently low and stable EMR values, ranging from 0.13 to 0.20, irrespective of the split ratio. In contrast, the \textit{Qwen2.5-1.5B-Instruct}~\cite{yang2024qwen2} and \textit{Qwen2.5-3B-Instruct}~\cite{yang2024qwen2} models exhibited an EMR consistently at 0.00 across all tested scenarios, indicating robust performance independent of data partitioning.

These findings underscore the robustness of the proposed method, demonstrating its capacity to maintain stable performance across diverse data distributions and varying sample sizes. This stability is particularly valuable in mitigating the risks associated with model overfitting or underfitting, enhancing the method's generalization capabilities across different experimental conditions.

\input{Table/EMR_at_various_split_ratios}

\subsubsection{\textit{AUROC Analysis}}

As presented in Table~\ref{tb:auroc}, the Area Under the Receiver Operating Characteristic (AUROC) results for the MedMCQA, MedQA, and MMLU datasets indicate a high degree of alignment between the frequency-based method and the logit-based approach. The AUROC values for both methods remain closely matched across all tested models and datasets, consistently achieving relatively high performance. This strong correlation suggests that the frequency-based method serves as a viable approximation of the logit-based approach, particularly in scenarios where logits are unavailable.

The minimal performance discrepancy between the two methods highlights the robustness and practicality of employing the frequency method as a substitute for logit-based evaluations. This is particularly advantageous in applications with restricted access to logit data, enabling a reliable and effective alternative for model performance assessment under such constraints.

\input{Table/Auroc}

\subsubsection{\textit{Reliability Measurement}}

A comparative analysis was conducted to evaluate model reliability using two distinct measurement strategies: frequency-based metrics and logit-based metrics, explicitly focusing on the MedMCQA dataset. The assessment centered on two critical performance indicators: Empirical Error Rate (EMR) and Average Prediction Set Size (APSS).

\begin{figure*}[h]
    \centering
    \includegraphics[width=\linewidth]{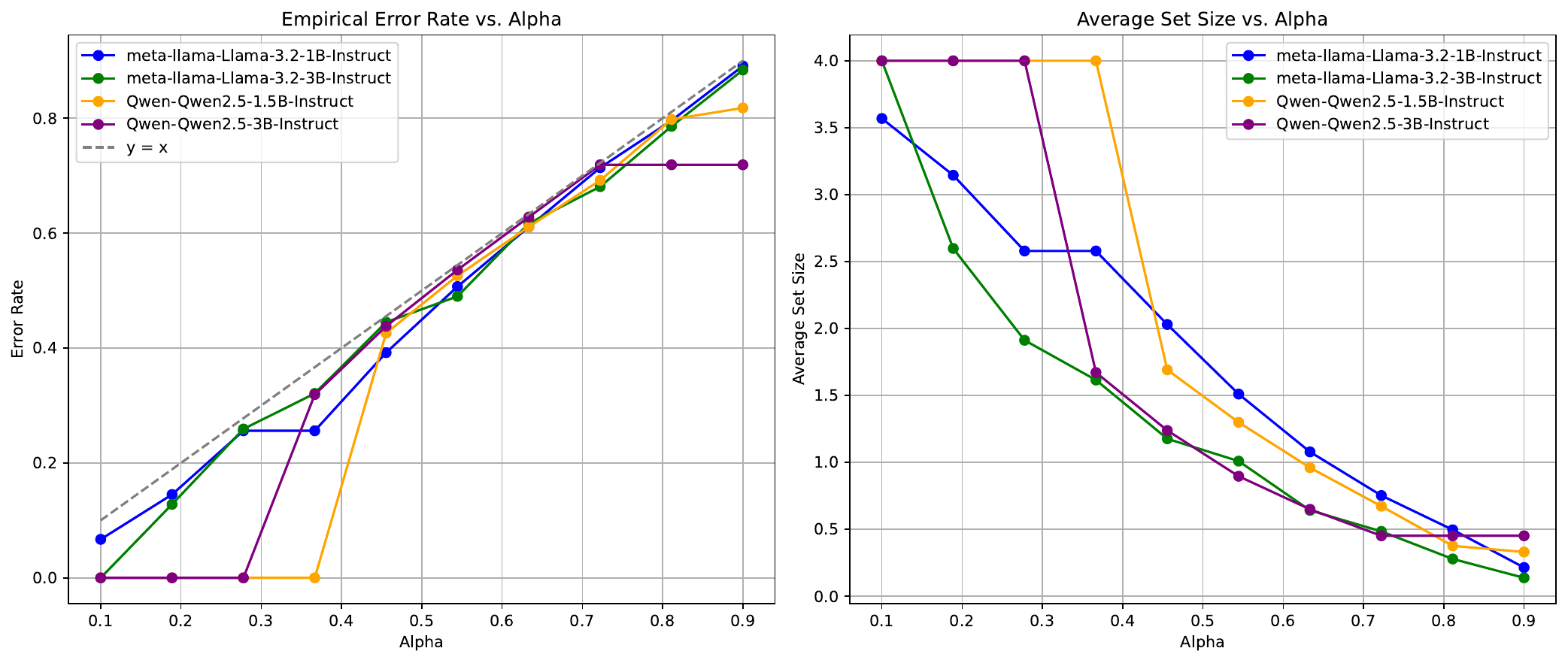}
    \caption{Reliability measurement using frequency-based metrics on the MedMCQA dataset, showing variability in EMR and APSS across different \( \alpha \) values.}
    \label{fig:frequency_emr_apss}
\end{figure*}

\begin{figure*}[h]
    \centering
    \includegraphics[width=\linewidth]{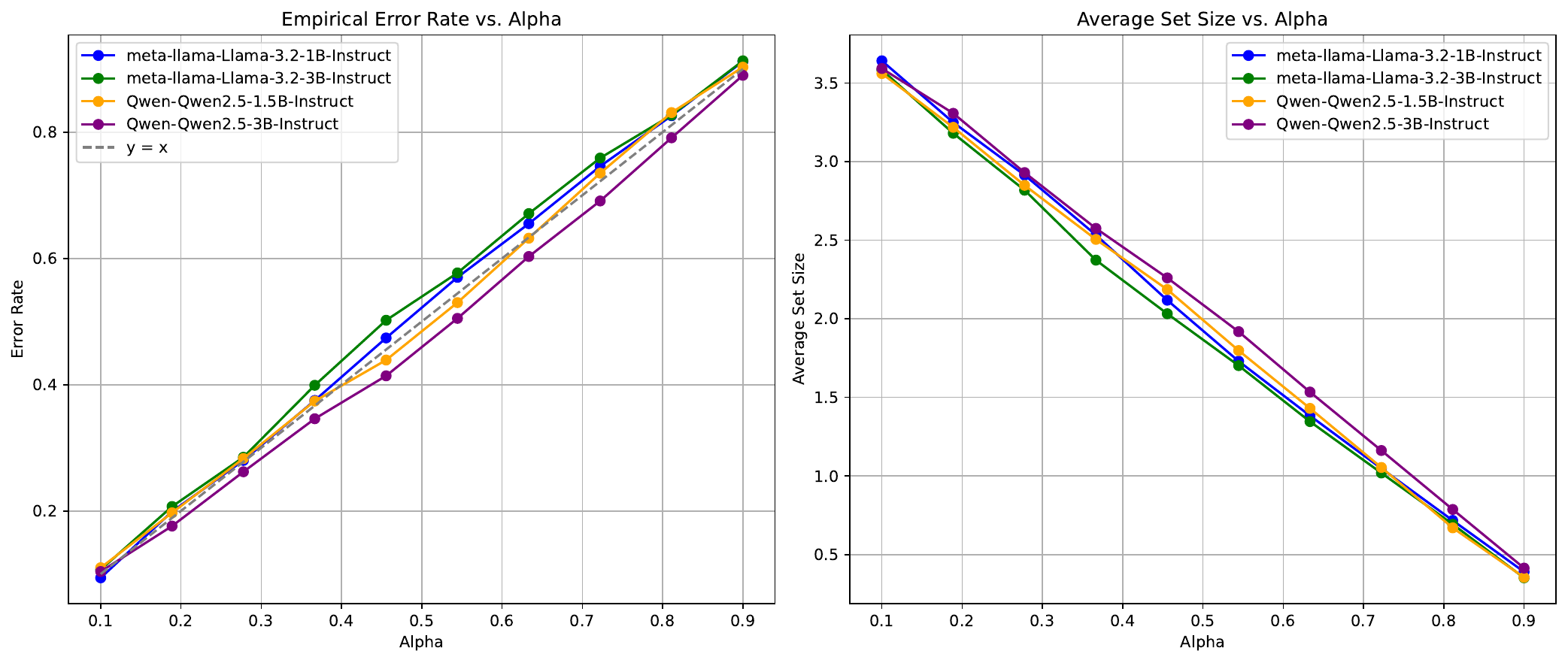}
    \caption{Reliability measurement using logit-based metrics on the MedMCQA dataset, demonstrates stable performance with a smooth increase in EMR and APSS as \( \alpha \) increases.}
    \label{fig:logit_emr_apss}
\end{figure*}

Figure~\ref{fig:frequency_emr_apss} presents the performance using frequency-based measurements, while Figure~\ref{fig:logit_emr_apss} illustrates the results with logit-based metrics. The logit-based approach demonstrates a consistent and gradual increase in EMR and APSS at varying risk levels \( \alpha \). values. The proposed models evaluated with logit-based metrics exhibit more stable performance, with error rates and prediction set sizes increasing smoothly and predictably as \( \alpha \) rises.

Conversely, the frequency-based approach, as shown in Figure~\ref{fig:frequency_emr_apss}, results in more pronounced fluctuations, particularly in EMR. The frequency-based method shows a steeper decline in average set size; however, the associated increase in EMR is more erratic when compared to the logit-based approach. It is indicated that frequency-based metrics may lead to greater variability in model performance as the confidence threshold changes.

The analysis suggests that the logit-based approach is more effective and reliable in maintaining consistent performance, particularly in managing error rates and prediction set sizes. It provides a more predictable balance between error rate and average set size, making it a preferred choice for high-stakes applications where reliability and stability are critical.

Nevertheless, frequency-based metrics offer a practical alternative in practical scenarios where access to the model's internal logits may not be feasible. While these metrics may not achieve the same level of precision and stability as the logit-based approach, they still provide a reasonable approximation of performance and reliability. This adaptability is advantageous in real-world applications where the internal model structure or logit data is inaccessible, enabling robust reliability assessments using only observable model outputs.

\section{Conclusions}

This study presents an enhanced CP framework to achieve risk control in medical MCQA tasks. By integrating self-consistency theory, the enhanced CP framework provides a statistically rigorous and interpretable method to improve the trustworthiness of LLMs in high-stakes medical applications by mitigating risks linked to model hallucinations. Comparative experiments demonstrated that the enhanced CP framework not only preserves CP's model-agnostic and distribution-free advantages extends its functionality beyond simple coverage control by introducing a monotonically decreasing loss function that effectively manages task-specific metrics. Additionally, the enhanced CP framework consistently achieved user-specified miscoverage rates. The results analysis also indicated a clear inverse relationship between the APSS and the associated risk level, highlighting the potential of APSS as a robust metric for quantifying LLM uncertainty. Future research will investigate the adaptation of this framework to other domain-specific QA tasks and its integration with advanced LLM architectures to improve reliability and transparency further.

\vspace{6pt} 

\bibliographystyle{unsrt}  
\bibliography{references}  

\end{document}

%% file: Table/APSS_results.tex
\begin{table}[H]
\caption{Results of the APSS metric at various risk levels.\label{tb:apss_results}}
\centering
\begin{tabularx}{\textwidth}{l l *{9}{X}}
\toprule
\textbf{Datasets} & \textbf{LLMs / $\alpha$} & \textbf{0.1} & \textbf{0.2} & \textbf{0.3} & \textbf{0.4} & \textbf{0.5} & \textbf{0.6} & \textbf{0.7} & \textbf{0.8} & \textbf{0.9} \\
\midrule

\multirow{4}{*}{MedMCQA} 
    & Llama-3.2-1B-Instruct & 3.57 & 3.15 & 2.58 & 2.03 & 1.51 & 1.08 & 1.08 & 0.50 & 0.22 \\
    & Llama-3.2-3B-Instruct & 4.00 & 2.60 & 1.91 & 1.37 & 1.01 & 0.75 & 0.48 & 0.28 & 0.12 \\
    & Qwen2.5-1.5B-Instruct & 4.00 & 4.00 & 4.00 & 1.98 & 1.54 & 1.11 & 0.74 & 0.44 & 0.33 \\
    & Qwen2.5-3B-Instruct & 4.00 & 4.00 & 4.00 & 1.45 & 1.17 & 0.75 & 0.45 & 0.45 & 0.45 \\
\midrule

\multirow{4}{*}{MedQA} 
    & Llama-3.2-1B-Instruct & 4.00 & 4.00 & 3.35 & 2.74 & 2.12 & 1.57 & 1.09 & 0.77 & 0.24 \\
    & Llama-3.2-3B-Instruct & 4.00 & 4.00 & 1.94 & 1.42 & 1.06 & 0.67 & 0.44 & 0.29 & 0.21 \\
    & Qwen2.5-1.5B-Instruct & 4.00 & 4.00 & 4.00 & 4.00 & 4.00 & 1.30 & 0.83 & 0.47 & 0.38 \\
    & Qwen2.5-3B-Instruct & 4.00 & 4.00 & 4.00 & 4.00 & 4.00 & 0.91 & 0.64 & 0.64 & 0.64 \\
\midrule

\multirow{4}{*}{MMLU} 
    & Llama-3.2-1B-Instruct & 3.89 & 3.32 & 2.83 & 2.14 & 1.53 & 1.04 & 0.68 & 0.21 & 0.03 \\
    & Llama-3.2-3B-Instruct & 2.43 & 1.66 & 1.24 & 0.92 & 0.65 & 0.49 & 0.25 & 0.25 & 0.25 \\
    & Qwen2.5-1.5B-Instruct & 4.00 & 4.00 & 1.41 & 1.06 & 0.79 & 0.54 & 0.43 & 0.43 & 0.43 \\
    & Qwen2.5-3B-Instruct & 4.00 & 4.00 & 1.06 & 0.83 & 0.83 & 0.83 & 0.83 & 0.83 & 0.83 \\

\bottomrule
\end{tabularx}
\end{table}

%% file: Table/EMR_at_various_split_ratios.tex
\begin{table}[H]
\caption{Exact Match Ratio (EMR) at various split ratios.\label{tb:EMR_at_various_split_ratios}}
\centering
\small
\renewcommand{\arraystretch}{1.2} 
\begin{tabularx}{\textwidth}{l l C C C C}
\toprule
\textbf{Datasets} & \textbf{LLMs / $\alpha=0.1$} & \textbf{Split Ratio = 0.1} & \textbf{0.3} & \textbf{0.5} & \textbf{0.7} \\
\midrule
\multirow{4}{*}{MedMCQA} 
    & Llama-3.2-1B-Instruct & 0.14 & 0.14 & 0.14 & 0.14 \\
    & Llama-3.2-3B-Instruct & 0.13 & 0.13 & 0.20 & 0.20 \\
    & Qwen2.5-1.5B-Instruct & 0.00 & 0.00 & 0.00 & 0.00 \\
    & Qwen2.5-3B-Instruct & 0.00 & 0.00 & 0.00 & 0.00 \\
\bottomrule
\end{tabularx}
\end{table}

%% file: Table/Auroc.tex
\begin{table}[H]
\caption{Results of the AUROC metric.\label{tb:auroc}}
\centering
\small
\renewcommand{\arraystretch}{1.2} 
\begin{tabularx}{\textwidth}{l l C C} 
\toprule
\textbf{Datasets} & \textbf{LLMs} & \textbf{Frequency} & \textbf{Logit} \\
\midrule

\multirow{4}{*}{MedMCQA} 
    & Llama-3.2-1B-Instruct & 0.5872 & 0.6197 \\
    & Llama-3.2-3B-Instruct & 0.6982 & 0.7005 \\
    & Qwen2.5-1.5B-Instruct & 0.6337 & 0.6517 \\
    & Qwen2.5-3B-Instruct & 0.6418 & 0.6930 \\
\midrule

\multirow{4}{*}{MedQA} 
    & Llama-3.2-1B-Instruct & 0.5819 & 0.6041 \\
    & Llama-3.2-3B-Instruct & 0.7375 & 0.7484 \\
    & Qwen2.5-1.5B-Instruct & 0.5752 & 0.6241 \\
    & Qwen2.5-3B-Instruct & 0.5693 & 0.6609 \\
\midrule

\multirow{4}{*}{MMLU} 
    & Llama-3.2-1B-Instruct & 0.6380 & 0.6358 \\
    & Llama-3.2-3B-Instruct & 0.7501 & 0.7657 \\
    & Qwen2.5-1.5B-Instruct & 0.7159 & 0.7771 \\
    & Qwen2.5-3B-Instruct & 0.6207 & 0.7784 \\
\bottomrule
\end{tabularx}
\end{table}